\documentclass{article}

\usepackage[preprint]{neurips_2023}

\usepackage[utf8]{inputenc} % allow utf-8 input
\usepackage[T1]{fontenc}    % use 8-bit T1 fonts
\usepackage{hyperref}       % hyperlinks
\usepackage{url}            % simple URL typesetting
\usepackage{booktabs}       % professional-quality tables
\usepackage{amsfonts}       % blackboard math symbols
\usepackage{nicefrac}       % compact symbols for 1/2, etc.
\usepackage{microtype}      % microtypography
\usepackage{xcolor}         % colors
\usepackage{graphicx}
\usepackage{amsmath}
\usepackage{amssymb}
\usepackage{multirow}
\usepackage{algorithm}
\usepackage{algorithmic}
\usepackage{diagbox}
\usepackage{natbib}
\setcitestyle{numbers,square}

\title{Improving Stability and Performance of Spiking Neural Networks through Enhancing Temporal Consistency}

\author{
{Dongcheng Zhao$^{1}$\footnotemark[1], \ Guobin Shen$^{1, 3}$\footnotemark[1], \ Yiting Dong$^{1, 3}$, \ Yang Li$^{1, 4}$, \ Yi Zeng$^{1, 2, 3, 4}$}\footnotemark[2]\\
$^1$ Brain-inspired Cognitive Intelligence Lab, Institute of Automation, Chinese Academy of Sciences\\ 
$^2$ Center for Excellence in Brain Science and Intelligence Technology, CAS\\
$^3$ School of Future Technology, University of Chinese Academy of Sciences \\
$^4$ School of Artificial Intelligence, University of Chinese Academy of Sciences \\
\texttt{\{zhaodongcheng2016, shenguobin2021,} \\ 
\texttt{dongyiting2020, liyang2019, yi.zeng\}@ia.ac.cn}
}

\begin{document}

\maketitle

\footnotetext[1]{These authors contributed equally.}
\footnotetext[2]{Corresponding Author.}
\begin{abstract}
Spiking neural networks have gained significant attention due to their brain-like information processing capabilities. The use of surrogate gradients has made it possible to train spiking neural networks with backpropagation, leading to impressive performance in various tasks. However, spiking neural networks trained with backpropagation typically approximate actual labels using the average output, often necessitating a larger simulation timestep to enhance the network's performance. This delay constraint poses a challenge to the further advancement of SNNs. Current training algorithms tend to overlook the differences in output distribution at various timesteps. Particularly for neuromorphic datasets, inputs at different timesteps can cause inconsistencies in output distribution, leading to a significant deviation from the optimal direction when combining optimization directions from different moments. To tackle this issue, we have designed a method to enhance the temporal consistency of outputs at different timesteps. We have conducted experiments on static datasets such as CIFAR10, CIFAR100, and ImageNet. The results demonstrate that our algorithm can achieve comparable performance to other optimal SNN algorithms. Notably, our algorithm has achieved state-of-the-art performance on neuromorphic datasets DVS-CIFAR10 and N-Caltech101, and can achieve superior performance in the test phase with timestep T=1.
%outperforms on neuromorphic datasets like DVS-CIFAR10 and N-Caltech101. Importantly, by enhancing temporal consistency, our algorithm can achieve superior performance during the testing phase with T=1. 
\end{abstract}

\section{Introduction}
Spike neural networks (SNNs), inspired by biological neural systems, exhibit unique advantages in processing spatiotemporal data. SNNs represent and transmit information through sparse, discrete spike sequences, which not only improves energy efficiency but also allows for better integration with neuromorphic chips~\cite{roy2019towards}, attracting the attention of researchers from various fields~\cite{zeng2022braincog}. However, the binary method of information transmission, which is similar to that of the human brain, possesses non-differentiable characteristics. This makes it challenging to directly apply the backpropagation algorithm to the training of Spiking Neural Networks (SNNs), consequently posing a significant obstacle in training deep SNNs.

Some researchers have attempted to incorporate brain-inspired learning principles into the modeling process of spiking neural networks, such as spike timing dependent plasticity (STDP)~\cite{dong2022unsupervised}  and short-term plasticity~\cite{zheng2021high}. Although these methods have addressed the training issues of SNNs to some extent, they still have limitations in dealing with more complex structures and tasks. Currently, there are two main strategies for achieving high-performance deep SNNs: one is the conversion-based approach~\cite{han2020rmp,wang2022signed,liu2022spikeconverter,li2022efficient}, and the other is the backpropagation (BP)-based approach~\cite{guo2022loss,li2021differentiable,dengtemporal,duan2022temporal}. The conversion-based method uses the activations of a pre-trained artificial neural network (ANN) as the firing rate of the SNN, allowing for a nearly lossless conversion from an ANN to an SNN. However, since this method requires firing rates to simulate activation values in ANNs accurately, it necessitates longer simulation steps and cannot fully exploit the spatiotemporal information processing capabilities of SNNs. On the other hand, the backpropagation-based training algorithm uses surrogate functions to approximate the gradients of the spike firing function, making it possible to train deep SNNs. Nevertheless, this approach also has limitations, such as gradient vanishing or exploding when dealing with more complex structures.

In contrast to ANNs that output results directly, SNNs need to accumulate outputs over multiple timesteps to make accurate judgments. However, this requires SNNs to use longer simulation steps to model more precise information, resulting in  greater computational resources when deploying on edge devices. The latency caused by large simulation length has also become an important factor hindering the development of SNNs. When SNNs receive inputs at different timesteps, especially for neuromorphic data with rich spatiotemporal characteristics, different output patterns are generated at different timesteps. Currently, the training of spiking neural networks based on the backpropagation algorithm mainly optimizes the difference between the average membrane potential or spike firing rate of the output layer and the accurate labels. Then, it uses the output at different timesteps to guide network training in the backward process, which does not consider the difference in output distribution under different timesteps. During training, the combined optimization direction of inconsistent outputs at different moments conflicts with the optimization direction of the average output value, thus preventing the network from being effectively optimized. During testing, the variance in the distribution at different moments can result in the network's test results being offset by incorrect inference results at certain moments, thereby reducing overall performance. Therefore, this paper proposes a new method to enhance the temporal consistency (ETC) at different timesteps during the training process, making the training of spiking neural networks more stable. At the same time, we have verified the method on multiple datasets, and the results show that the method has achieved outstanding performance on all datasets. The main contributions of this paper are as follows:
\begin{itemize}
    \item Through theoretical analysis, this paper reveals the limitations of existing backpropagation methods in the training of SNNs, especially in dealing with the differences in output distribution between different moments.
    \item We propose a novel method that enhances the temporal consistency across different moments, which improves the stability of SNN training and significantly reduces the required timesteps.
    \item To validate the superiority of our proposed method, we have conducted experiments on the static datasets CIFAR10, CIFAR100, and ImageNet and the neuromorphic datasets DVS-CIFAR10 and N-Caltech101. The results show that our algorithm achieves the best performance on neuromorphic datasets and delivers competitive performance compared to other state-of-the-art algorithms on static datasets.
\end{itemize}

\section{Related Work}
Researchers mainly improve the BP-based SNNs from the following two aspects. First, they tried to enhance the information transmission capability of spiking neural networks. A series of studies~\cite{fang2021incorporating,dingbiologically,yaoglif,yin2021accurate,wang2022ltmd} focused on introducing learnable parameters, such as membrane potential constants and thresholds, into spiking neurons to construct different types of spiking neurons. This approach  enhances the adaptability of neural networks to various information types. NeuNorm~\cite{wu2019direct}, tdBN~\cite{zheng2021going}, and TEBN~\cite{duan2022temporal} incorporated various normalization techniques to enhance information transmission in deep spiking neural networks. These approaches effectively mitigate gradient vanishing and explosion issues, improving network performance and stability. 
In the works by ~\cite{yao2021temporal, zhu2022tcja, yao2023attention}, various attention mechanisms were introduced across different dimensions to guide SNNs in transmitting more task-relevant information, as a result of this enhancing their overall performance and adaptability.

Another group of researchers attempted to improve the structure of SNNs. Drawing inspiration from brain-inspired structures, LISNN~\cite{cheng2020lisnn} introduced lateral connections, while BackEISNN~\cite{zhao2022backeisnn} incorporated self-feedback connections to enhance SNN's information processing capabilities. Furthermore, SewResNet~\cite{fang2021deep} designed a more suitable residual module for SNNs, providing a simple method for training deep SNNs. Spikformer~\cite{zhou2022spikformer} combined self attention with SNNs, constructing a high-performance SNN structure. AutoSNN~\cite{na2022autosnn} and NASSNN~\cite{kim2022neural} employed the neural architecture search technique to design more optimal structures for SNNs.

However, the works above all used the average membrane potential or spike firing rate at the output layer for prediction without considering the impact of the output distribution of SNNs at different timesteps on performance. Reducing the differences in SNN output distribution at various timesteps is crucial for constructing a stable and high-performance SNNs.
\begin{figure}
    \centering
    \includegraphics[width=1.0\linewidth]{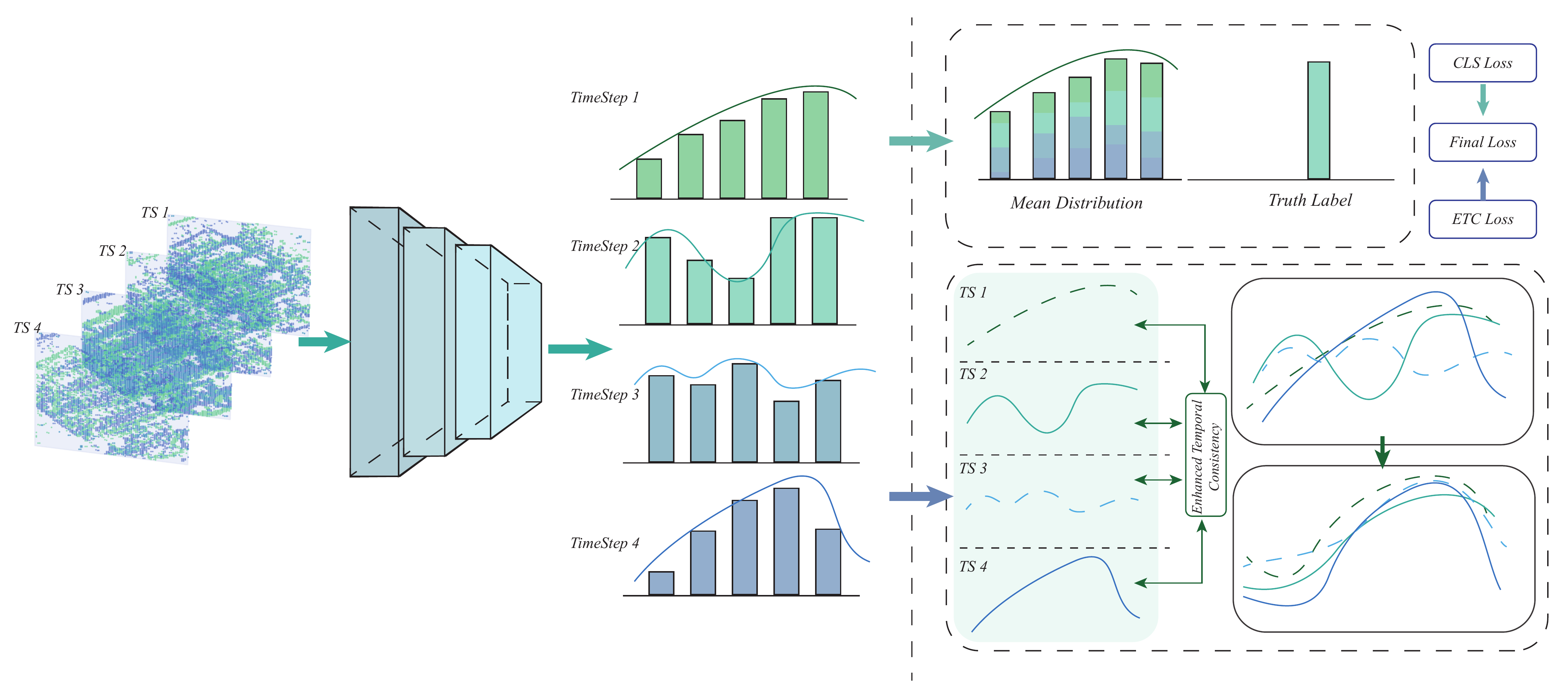}
    \caption{The whole training pipeline of our model. SNNs receive input data at various timesteps and generate corresponding outputs at each timestep. The ETC constraint helps ensure output distribution consistency across different timesteps.   }
    \label{pelu_fig}
\end{figure}

\section{Method}
In this section, we first introduce the spiking neurons used and theoretically analyze the issue of uneven distribution at different timesteps. Ultimately, we introduce the enhancing temporal consistency constraint to standardize the output distribution at different timesteps.
\subsection{Spiking Neuron Model}
The leaky integrate-and-fire (LIF) neuron, as the most commonly used neuron model in deep spiking neural networks, describes the complex dynamics of biological neurons with a relatively simple differential equation as shown in Eq.~\ref{lif}:
\begin{equation}
\begin{aligned}
    \tau_m \frac{dV}{dt} &= -V + RI \quad V \le V_{th} \\
    S &= H(V - V_{th})
\end{aligned}
\label{lif}
\end{equation}
$V$ represents the membrane potential, $\tau_m = RC$ is the membrane potential time constant, $I = \sum_j W_{ij}S_j$ denotes the input current obtained by aggregating presynaptic spikes, $W$ denotes the connection weight from pre-synaptic to post-synaptic neurons, and $H$ refers to the step function for spike emission. When the membrane potential surpasses the threshold $V_{th}$, the neuron emits a spike $S$ and resets to the resting potential $V_r$. In this study, we set $V_{th}$ to 0.5, $\tau_m$ to 2, $R$ to 1, and the resting potential $V_r$ to 0. By using the first-order Euler method, we obtain the discretized representation of the above differential equation as shown in Eq.~\ref{lif_dis}: 
\begin{equation}
     V_{t+1} = (1 -  \frac{1}{\tau_m})V_t + \frac{1}{\tau_m}I_t
    \label{lif_dis}
\end{equation}

In order to use the backpropagation algorithm for network training, we employ surrogate gradients to approximate the gradients of the spike firing function, as follows:
\begin{equation}
\label{eq6}
\frac{\partial H}{\partial V_t} =\left\{
\begin{aligned}
& 0 , & |V_t-V_{th}| > \frac{1}{a} \\
& -a^2|V_t - V_{th}| + a  ,& |V_t-V_{th}| \leq \frac{1}{a}
\end{aligned}
\right.
\end{equation}

$a$ is a hyperparameter used to control the shape of the surrogate gradient. In this study, we set $a$  to 2.

\subsection{The Training and Inference Procedures compared SNNs with ANNs}
Unlike traditional artificial neural networks, SNNs adjust their weights during training based on outputs at multiple timesteps. During the inference of the network, the final output is calculated by weighting the outputs at various timesteps. For deep SNNs, the most widely used approach is to employ the average membrane potential of the neurons in the last layer as the final output, resulting in more efficient and accurate inference. 

We represent the average membrane potential of the last layer as $O_\text{mean} = \frac{1}{T} \sum_{t=1}^T V_t$, where T is the total number of timesteps and $V_t$ is the membrane potential at timestep $t$. The softmax function is applied to convert the average membrane potential into output probabilities   $P_\text{mean}= \text{softmax}(O_\text{mean})$. During the network training process, we use the cross-entropy loss to minimize the difference between the actual outputs $P_\text{mean}$ and expected outputs $y$ as shown in Eq.~\ref{loss_ce}.  
\begin{equation}
L_{CE} = -\sum_{i=1}^C y_i \cdot \log(P_\text{mean})	
 %L_{CE} = -\sum_{i=1}^n y_i log(p_i)
 \label{loss_ce}
\end{equation}
C is the number of categories.  By applying the chain rule, we can compute the gradient of the loss function concerning the network parameters $W$. 
\begin{equation}
    \begin{aligned}
        \frac{\partial L_{CE}}{\partial W} = \frac{\partial L_{CE}}{\partial O_\text{mean}}\frac{\partial O_\text{mean}}{\partial V_t} \frac{\partial V_t}{\partial W}=\frac{1}{T}\sum_{t=1}^T[P_\text{mean} - y] \frac{\partial V_t}{\partial W}
    \end{aligned}
    \label{loss_snn}
\end{equation} 
For ANNs, we use $O$ to denote the final output and $P$ to represent the probability distribution after the softmax operation. Using the same loss function, we can also obtain the gradient of the loss concerning the network parameters:
\begin{equation}
    \begin{aligned}
        \frac{\partial L_{CE}}{\partial W} = \frac{\partial L_{CE}}{\partial O}\frac{\partial O}{\partial W} =[P_\text{mean} - y] \frac{\partial O}{\partial W}
    \end{aligned}
    \label{loss_ann}
\end{equation} 

\begin{figure}
    \centering
    \includegraphics[width=1.0\linewidth]{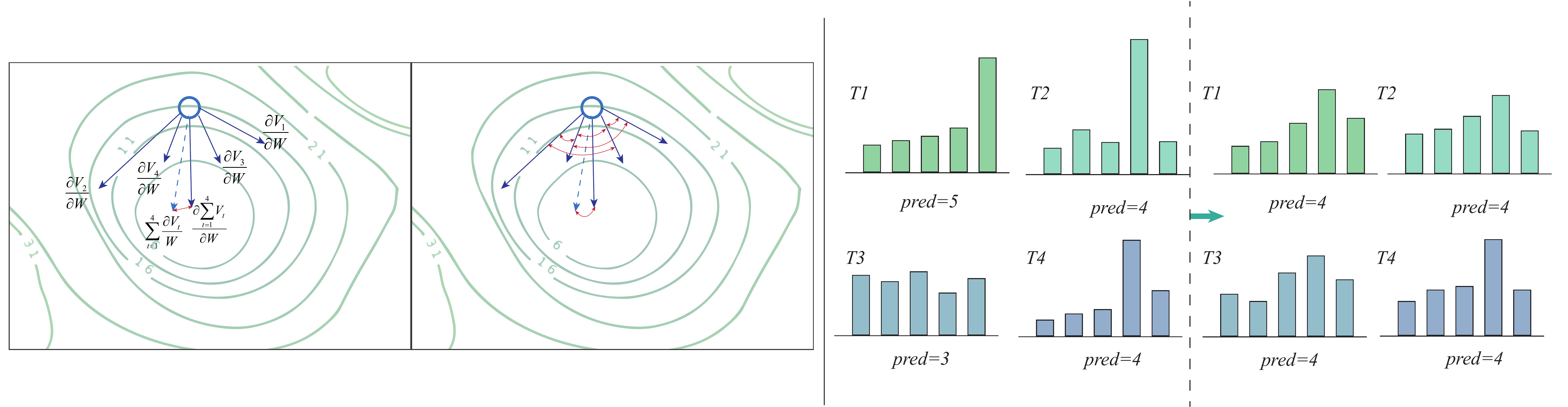}
    \caption{The training and inference procedure compared our method with the traditional methods. For traditional methods, consistency in the distribution across different timesteps can cause discrepancies in the optimization direction and lead to misjudgments during inference.}
    \label{ann_SNN}
\end{figure}

As shown in Eq.~\ref{loss_snn} and Eq.~\ref{loss_ann}, the final output determines the partial derivative of the weight. In minimizing the loss, the output $O$ of the ANN will gradually approach the actual label. At this time, using the partial derivative of $O$ concerning W will more accurately control the optimization direction of the ANN. However, the optimization direction of the weights in SNN depends on the direction of the partial derivative of the output $V_t$ concerning the weight at each moment. SNN optimizes the distance between the average membrane potential and the actual label. As shown in Fig.~\ref{ann_SNN}, when the optimization directions at different moments are inconsistent or even significantly different, there will be a severe mismatch between $\sum_{t=1}^T\frac{\partial V_t}{\partial W} $ and $ \frac{\partial \sum_{t=1}^TV_t}{\partial W} $, which significantly interferes with the optimization direction of the spiking neural network. Simultaneously, as shown on the right side of Fig.~\ref{ann_SNN}, during the inference phase of the network, inconsistencies in the distribution at different timesteps can lead to overall network results being skewed by erroneous results, thereby leading to a decline in performance.

\subsection{Reducing Temporal Diversity Procedure}
As discussed above, the performance degradation of SNNs is due to the mismatch of output distributions at different timesteps. In order to enhance the consistency between different timesteps, we propose an enhancing temporal consistency constraint, aiming to make the distributions at each timestep as similar as possible. First, we define the output probability distribution at each timestep $P_t^i$  as shown in Eq.~\ref{eq_dist}:
\begin{equation}
    \begin{aligned}
        P_t^i(V_t;\tau) = softmax(V_t^i;\tau) = \frac{exp(V_t^i/\tau)}{\sum_j exp(V_t^j/\tau)}
    \end{aligned}
    \label{eq_dist}
\end{equation} 

The temperature parameter $\tau$ controls the smoothness of the model's output distribution, which is more conducive to learning the relationships between different categories~\cite{hinton2015distilling}. Here we set $\tau=4$. After obtaining the output distributions at different timesteps, we aim to minimize the distribution gap between the output $P_t$ at time t and the outputs at other timesteps, here we use the Kullback-Leibler (KL) divergence. The loss function is shown in Eq.~\ref{loss_t}.
\begin{equation}
    \begin{aligned}
        L^t_{ETC} &= \frac{1}{T-1}\sum_{m=1,m\neq t}^{T}KL(P_m|| P_t) = \frac{1}{T-1}\sum_{m=1,m\neq t}^{T} \sum_{i=1}^{C}P_m^i log\frac{P_m^i}{P_t^i}\\
        &=\frac{1}{T-1}\sum_{m=1,m\neq t}^{T} \sum_{i=1}^{C}(P_m^i log P_m^i - P_m^i log P_t^i)
    \end{aligned}
    \label{loss_t}
\end{equation} 

To avoid model collapse, we do not propagate the gradient through $P_m$. As a result, the final loss can be represented as shown in Eq.~\ref{loss_all}:
\begin{equation}
    \begin{aligned}
        L^t_{ETC} = -\frac{1}{T-1}\sum_{m=1,m\neq t}^{T} \sum_{i=1}^{C} P_m^i log P_t^i \Rightarrow L_{RTD} = -\frac{1}{T}\frac{1}{T-1}\sum_{t=1}^T\sum_{m=1,m\neq t}^{T} \sum_{i=1}^{C} P_m^i log P_t^i
    \end{aligned}
    \label{loss_all}
\end{equation} 

Thus, the final loss function can be written as a dynamic combination of cross-entropy loss and ETC loss, as shown in Eq.~\ref{loss3}.
\begin{equation}
    \begin{aligned}
        L_{all} = L_{CE} + \lambda \tau^2 L_{RTD} 
    \end{aligned}
    \label{loss3}
\end{equation} 

$\lambda$ is the weight constraint term to control the influence of ETC loss on overall loss. Here we set $\lambda=1$. We can obtain the partial derivative of the total loss concerning the weights $\frac{\partial L_{all}}{\partial W}$:
\begin{equation}
    \begin{aligned}
        \frac{\partial L_{all}}{\partial W} &= \frac{\partial L_{CE}}{\partial W} +  \lambda \tau^2 \frac{\partial L_{ETC}}{\partial W}\\&=\frac{1}{T}\sum_{t=1}^T[P_\text{mean} - y] \frac{\partial V_t}{\partial W} + \lambda \tau^2 \frac{1}{T}\frac{1}{T-1} \sum_{t=1}^T\sum_{m=1,m\neq t}^T(P_t-P_m) \frac{\partial V_t}{\partial W} 
    \end{aligned}
    \label{par_all}
\end{equation}

As shown in Eq.~\ref{par_all}, the first term ensures that the average output is close to the actual target, while the second term ensures the consistency of the output distribution at each timestep. Combining these two loss terms can make the output at each moment as accurate as possible, thus better guiding the optimization direction of the output for the weights at each moment. By using the output of different timesteps to approximate each other and correcting erroneous prediction moments, this combination of loss functions can also prevent overconfident predictions, thereby further improving the model's generalization ability on the test dataset. Moreover, this constraint can be considered a self-distillation process of the model. It uses the dark knowledge of the model's output at other timesteps to provide soft labels for each timestep, optimizing the output at each timestep and fully utilizing the temporal information of SNNs.

\section{Experiments}
In order to demonstrate the superiority of our algorithm, we conduct experiments on multiple datasets, including static datasets such as CIFAR-10~\cite{krizhevsky2009learning}, CIFAR-100~\cite{xu2015empirical}, and ImageNet~\cite{russakovsky2015imagenet}, as well as neuromorphic datasets like DVS-CIFAR10~\cite{li2017cifar10} and N-Caltech101~\cite{orchard2015converting}. We develop the  SNN code based on the open-source framework BrainCog with NVIDIA A100 graphic processing unit (GPU), employing the AdamW optimizer with a weight decay setting of 0.0001. The learning rate is set to 0.001 with the cosine annealing strategy. The batch size is set to 128. In our experiments, we set the total training epochs to 600. The experiments are repeated five times randomly. We have reported the mean and standard deviation of the corresponding performance.

\begin{table*}[t]
\centering
    \caption{Compare with existing works on static image datasets.}
    \label{static}
    % \label{sample-table}
        \begin{tabular}{ccccc}
            \hline
            \multicolumn{1}{c}{\bf Dataset}               & \multicolumn{1}{c}{\bf Model}              & {\bf Architecture} & \multicolumn{1}{c}{\bf Simulation Step} & \multicolumn{1}{c}{\bf Accuracy} \\
            \hline
            \multicolumn{1}{c}{\multirow{15}{*}{CIFAR10}}
            & Opt~\cite{bu2021optimal}     & ResNet-18    & 4       & 90.43        \\
            & CSTDB~\cite{rathi2019enabling}       & ResNet-20            & 250       & 92.22        \\
            & Diet-SNN~\cite{rathi2020diet}          & ResNet-20            & 10        & 92.54         \\
            %& ~\cite{wu2018spatio}              & STBP            & CIFARNet             & 12        & 89.83   \\
            & NeuNorm~\cite{wu2019direct}                  & CIFARNet             & 12        & 90.53   \\
            & TSSL-BP~\cite{zhang2020temporal}               & CIFARNet             & 5         & 91.41   \\
            & BPSTA~\cite{shen2022backpropagation}            & 7-layer-CNN          & 8         & 92.15   \\
            & NASSNN~\cite{kim2022neural}                       & NAS  & 5         & 92.73   \\
            & AutoSNN~\cite{na2022autosnn}                       & NAS  & 16        & 93.15   \\
            & tdBN~\cite{zheng2021going}               & ResNet-19            & 6         & 93.16   \\
            & PLIF~\cite{fang2021incorporating}   &PLIFNet & 8 &93.5\\
            & TET~\cite{dengtemporal}                       & ResNet-19            & 6         & 94.50   \\
            %&

             & GLIF~\cite{yaoglif}  & ResNet-19      & 6      & 95.03  \\
             &TKS~\cite{dong2023temporal}   & ResNet-19      & 4      & 95.3 \\
             & Rec-Dis~\cite{guo2022recdis} & ResNet-19  & 6         & 95.55   \\
            & TEBN~\cite{duan2022temporal}     & ResNet-19      & 6     & 95.60         \\
            \cline{2-5}
            & \multirow{4}{*}{\textbf{Our Method}}       & SEW-ResNet-18            & 2         & 94.65 $\pm$ 0.08 \\
            &                                             & SEW-ResNet-18            & 4         & 95.4 $\pm$ 0.07 \\
            &                                             & SEW-ResNet-18           & 6         & 95.73 $\pm$ 0.02 \\
            &                                            & SEW-ResNet-18            & 8         & 95.84 $\pm$ 0.03 \\
            \hline
            \multicolumn{1}{c}{\multirow{12}{*}{CIFAR100}}
            % & \cite{bu2021optimal}             & Conversion & ResNet-18            & 8       & 75.67        \\
            %& \cite{rathi2019enabling}      & Hybrid training & VGG-11               & 125       & 67.87  \\
            & Diet-SNN\cite{rathi2020diet}                  & ResNet-20            & 5         & 64.07  \\
            & BPSTA~\cite{shen2022backpropagation}             & ResNet34             & 8         & 69.32  \\
            & AutoSNN~\cite{na2022autosnn}                         & NAS  & 16        & 69.16  \\
            & NASSNN~\cite{kim2022neural}                        & NAS  & 5         & 73.04   \\
            & TET~\cite{dengtemporal}                    & ResNet-19            & 6         & 74.72  \\
            & Rec-Dis~\cite{guo2022recdis}                     & ResNet-19            & 4         & 74.10   \\
            & TKS~\cite{dong2023temporal}  & ResNet-19      & 4      & 76.2\\
            & GLIF~\cite{yaoglif}  & ResNet-19      & 6      & 77.35  \\
            & TEBN~\cite{duan2022temporal}  & ResNet-19      & 6     & 78.76        \\

            \cline{2-5}
            & \multirow{4}{*}{\textbf{Our Method}}         & SEW-ResNet-18            & 2         & 75.96 $\pm$ 0.24\\
            &                                             & SEW-ResNet-18            & 4         & 77.65 $\pm$ 0.13 \\
            &                                             & SEW-ResNet-18            & 6         & 78.25 $\pm$ 0.11 \\
            &                                             & SEW-ResNet-18            & 8         & 78.32 $\pm$ 0.07 \\
            \hline
            \multicolumn{1}{c}{\multirow{9}{*}{ImageNet}}
            %& \cite{bu2021optimal}             & Conversion & ResNet-34         & 16        & 59.35        \\
            %& \cite{rathi2019enabling}      & Hybrid training & ResNet-34            & 250       & 61.48              \\
          & tdBN~\cite{zheng2021going}  & Spiking-ResNet-34  & 6   & 63.72     \\
          %& \cite{fang2021incorporating} & PLIF &ResNet-34 & 7 &67.04\\
          & SEW~\cite{fang2021deep}& SEW-ResNet-34        & 4         & 67.04     \\
          & Rec-Dis~\cite{guo2022recdis} & ResNet-34  & 6         & 67.33   \\
            & TET~\cite{dengtemporal}          & SEW-ResNet-34        & 4         & 68.00      \\
            & TEBN~\cite{duan2022temporal}      & SEW-ResNet-34       & 4     & 68.28         \\
            & GLIF~\cite{yaoglif} & ResNet-34     & 6     & 69.09  \\
            & TKS~\cite{dong2023temporal}   & SEW-ResNet-34     & 4      & 69.6 \\
            \cline{2-5}
            & \multirow{3}{*}{\textbf{Our Method}}           & SEW-ResNet-18            & 4         & 63.70 \\
            &                                                & SEW-ResNet-34            & 4         & 68.54 \\
            &                                                & SEW-ResNet-34            & 6         & 69.64 \\
            \hline
            % \hline
        \end{tabular}
\end{table*}
As shown in Tab.~\ref{static}, we conduct experiments on static datasets based on SEW-ResNet and compare them with other state-of-the-art algorithms. We use the SEW-ResNet18 structure and perform experiments on the CIFAR10 and CIFAR100 datasets under simulation step lengths of 2, 4, 6, and 8, respectively. On the CIFAR10 dataset, our model achieves the accuracy of 95.73\% at the simulation step length of 6, surpassing all other algorithms and achieving the current best performance. For the CIFAR100 dataset, even with a relatively small network, we achieve performance comparable to the current best algorithm, only 0.4\% lower than the performance of TEBN. For the more complex ImageNet dataset, we conduct experiments based on SEW-ResNet-18 and SEW-ResNet-34. The results show that compared with the GLIF algorithm, the performance of our model has improved by 0.6\%. What's more, compared with the original SEW algorithm, our performance has improved by about 2\%. This fully demonstrates the superiority of our algorithm.
\subsection{Results on Neuromorphic Datasets}
Compared to static datasets, neuromorphic datasets reveal richer spatiotemporal features, thereby better highlighting the advantages of spiking neural networks. The DVS-CIFAR10 and N-Caltech101 datasets convert original image information into event information through dynamic vision sensors. In our study, we first resize the input samples to a fixed size of 48$\times$48 and adopt the VGGSNN structure used in TET. As shown in Tab.~\ref{neuromorphic}, our algorithm has achieved the best performance on both datasets. On the DVS-CIFAR10 dataset, our performance improved by 2\% compared to TET and by 0.8\% compared to TEBN. On the N-Caltech101 dataset, we has achieved the accuracy of 85.53\%, a 1.4\% improvement compared to TKS.
Meanwhile, our performance has significantly surpassed that of the EventMix and NDA algorithms, which use data augmentation techniques. Due to the characteristic of neuromorphic datasets where inputs at different times vary, the output distribution also changes. In such cases, introducing the method we propose in this paper can better standardize the membrane potential distribution at different timesteps, thereby significantly enhancing the network's performance.     
\begin{table*}[htbp]
\centering
    \caption{Compare with existing works on neuromorphic datasets. $^*$ indicates using a model with tdBN~\cite{zheng2021going}.}
    \label{neuromorphic}
    % \label{sample-table}
        \begin{tabular}{ccccc}
            \hline
            \multicolumn{1}{c}{\bf Dataset} & \multicolumn{1}{c}{\bf Model}      & \multicolumn{1}{c}{\bf Architecture} & \multicolumn{1}{c}{\bf Simulation Step} & \multicolumn{1}{c}{\bf Accuracy}   \\
           \hline
           \multicolumn{1}{c}{\multirow{9}{*}{DVS-CIFAR10}}
                & tdBN~\cite{zheng2021going}                    & ResNet-19     & 10    & 67.8    \\
                %& ~\cite{kugele2020efficient}   & Streaming Rollout    & DenseNet      & 10    & 66.8    \\
                & LIAF~\cite{wu2021liaf}                               & LIAF-Net      & 10    & 71.70   \\
                & LIAF~\cite{wu2021liaf}                                 & LIAF-Net      & 10    & 70.40   \\
                & AutoSNN~\cite{na2022autosnn}                              & NAS           & 16    & 72.50   \\
                & BPSTA~\cite{shen2022backpropagation}                 & 5-layer-CNN   & 16    & 78.95   \\
                &Rec-Dis~\cite{guo2022recdis}                          & ResNet-19     & 10    & 72.42   \\
                & TET~\cite{dengtemporal}                        & VGGSNN        & 10    & 83.17   \\
                & TEBN~\cite{duan2022temporal}     & VGGSNN     & 10     & 84.90         \\
                & TKS~\cite{dong2023temporal}     & VGGSNN$^*$      & 10      & 85.3 \\
                 \cline{2-5}
                & \multirow{2}{*}{\textbf{Our Method}}        & VGGSNN & 10    & 85.35 $\pm$ 0.40 \\
                &                                                   & VGGSNN$^*$   & 10    & 85.95 $\pm$ 0.43 \\
               \hline
              \multicolumn{1}{c}{\multirow{6}{*}{N-Caltech101}}
                & ConvertSNN~\cite{kugele2020efficient}                    & VGG11         & 20    & 55.0    \\
                & Dart~\cite{ramesh2019dart}                          & N/A           & N/A   & 66.8    \\
                & TCJA~\cite{zhu2022tcja}                   & TCJAnet         & 14    & 78.5   \\
                &NDA~\cite{li2022neuromorphic}      & ResNet-19&10&78.6 \\
                &EventMix~\cite{shen2022eventmix}  &    ResNet-18 & 10&79.5 \\
                &TKS~\cite{dong2023temporal}    & VGGSNN$^*$      & 10      & 84.1 \\
                \cline{2-5}
                & \multirow{2}{*}{\textbf{Our Method}}                         & VGGSNN          & 10    & 83.33 $\pm$ 0.41 \\
                &                                                               & VGGSNN$^*$          & 10    & 85.53 $\pm$ 0.09 \\
           \hline
        \end{tabular}
\end{table*}

%消融实验 
%画图分析/*
\subsection{Ablation Studies}
To verify the effectiveness of our algorithm, we conduct ablation experiments on the DVS-CIFAR10 and N-Caltech101 datasets. As shown in Tab.~\ref{aba}, without adding the ETC module, the accuracy of the N-Caltech101 dataset is only 78.28\%. However, after adding the ETC module with $\lambda=1$ and $\tau=4$, the network performance improves to 83.3\%, an increase of 5\%. Under the same settings, the performance of the DVS-CIFAR10 dataset is also improved by 2.5\%. It is worth noting that the hyperparameter $\tau$ controls the smoothness of the output at different moments. If $\tau$ is too large, the output will be too smooth to show significant differences.
Conversely, if $\tau$ is too small, the output will be overly confident and unable to reflect the relationships between different categories. Furthermore, the $\lambda$ parameter controls the influence of the ETC module on the final loss. If $\lambda$ is too large or too small, it cannot effectively guide the loss function to update the weights. To better illustrate the impact of these hyperparameters, we test the performance of the ETC algorithm under different hyperparameter settings.
\begin{table}[htbp]
  \centering
  \caption{Ablation analysis on ETC loss on DVS-CIFAR10 and N-Caltech101 datasets, as well as the sensitivity analysis of different hyperparameters}
    \begin{tabular}{ccccccccc}
          & \multicolumn{4}{c}{DVS-CIFAR10}      & \multicolumn{4}{c}{N-Caltech101} \\
       \midrule       $\lambda=0.$ & \multicolumn{4}{c}{82.9}             & \multicolumn{4}{c}{78.28}\\
         \midrule  \diagbox{$\tau$}{$\lambda$}  & 0.1   & 1     & 2     & 8      & 0.1   & 1     & 2     & 8 \\
    \midrule
    1     & 82.20  & 85.10  & 84.50  & 84.50   & 80.11  & 80.92 & 80.00  & 80.69\\
    2     & 85.70  & 85.90  & 84.60  & 84.40    & 81.84   & 82.13    & 80.57    & 80.00  \\
    4     & 85.30  & 85.40  & 85.60  & 85.30   & 81.72    & 83.33    & 81.72    & 81.60  \\
    8     & 85.60  & 85.00  & 85.10  & 84.40  & 81.95    & 82.03    & 80.92    & 80.34  \\
    16    & 85.40  & 85.00  & 85.30  & 84.60   & 81.38    & 81.95    & 81.95    & 81.03 \\
    \hline
    \end{tabular}%
  \label{aba}%
\end{table}%

\subsection{Temporal Consistency Verification}
\begin{figure}[h]
\centering
\includegraphics[width=0.8\columnwidth]{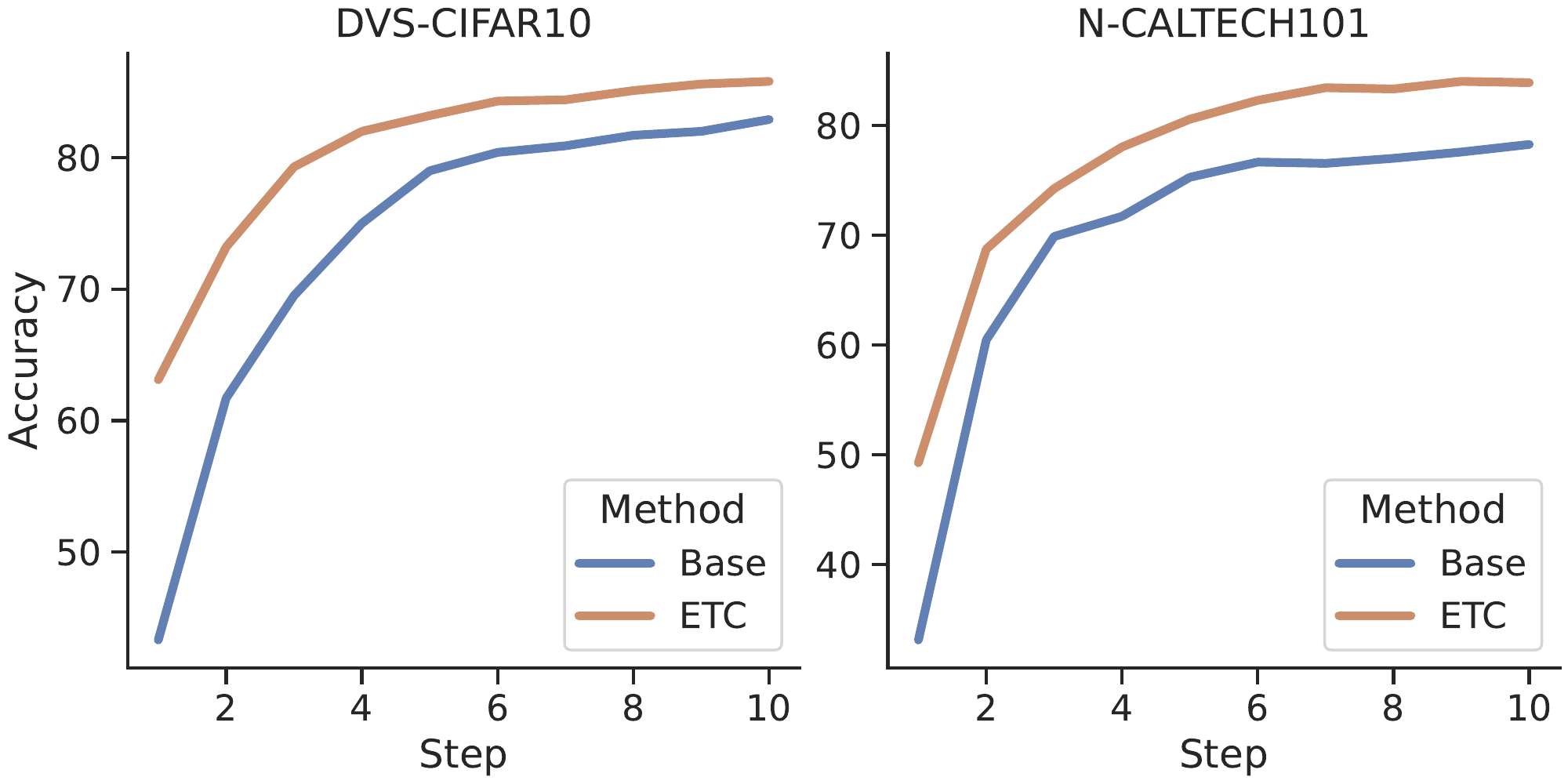}
\caption{Test Accuracy in different timesteps on DVS-CIFAR10 and N-Caltech101, the model is trained at timestep 10.}
\label{timefig}
 \end{figure}

In addition to performance, latency is a crucial factor that constrains the development of SNNs. When the output distribution at each moment becomes more consistent, we can achieve higher accuracy in the testing phase using only a shorter simulation length. We have validated the results on the DVS-CIFAR10 and N-Caltech101 datasets. During the training phase, we use the simulation timestep T=10, while in the testing phase, we conduct separate experiments for different timesteps. As shown in Fig.~\ref{timefig}, with a simulation step size of 1, the accuracy of the conventional algorithm is only 43.3\% for the DVS-CIFAR10 dataset.
In contrast, our ETC algorithm achieves an accuracy of 63.1\%, an improvement of about 20\%. At a step size of 5, the ETC algorithm already surpasses the performance of the traditional algorithm at a step size of 10, significantly reducing the network's latency. For the N-Caltech101 dataset, when the step size is 1, our algorithm has improved by 16\% compared to the baseline. At a step size of 5, the accuracy of the ETC algorithm has reached 80.57\%, which is 2\% higher than the baseline at a step size of 10.
 
\begin{figure}[h]
\centering
\includegraphics[width=1.0\columnwidth]{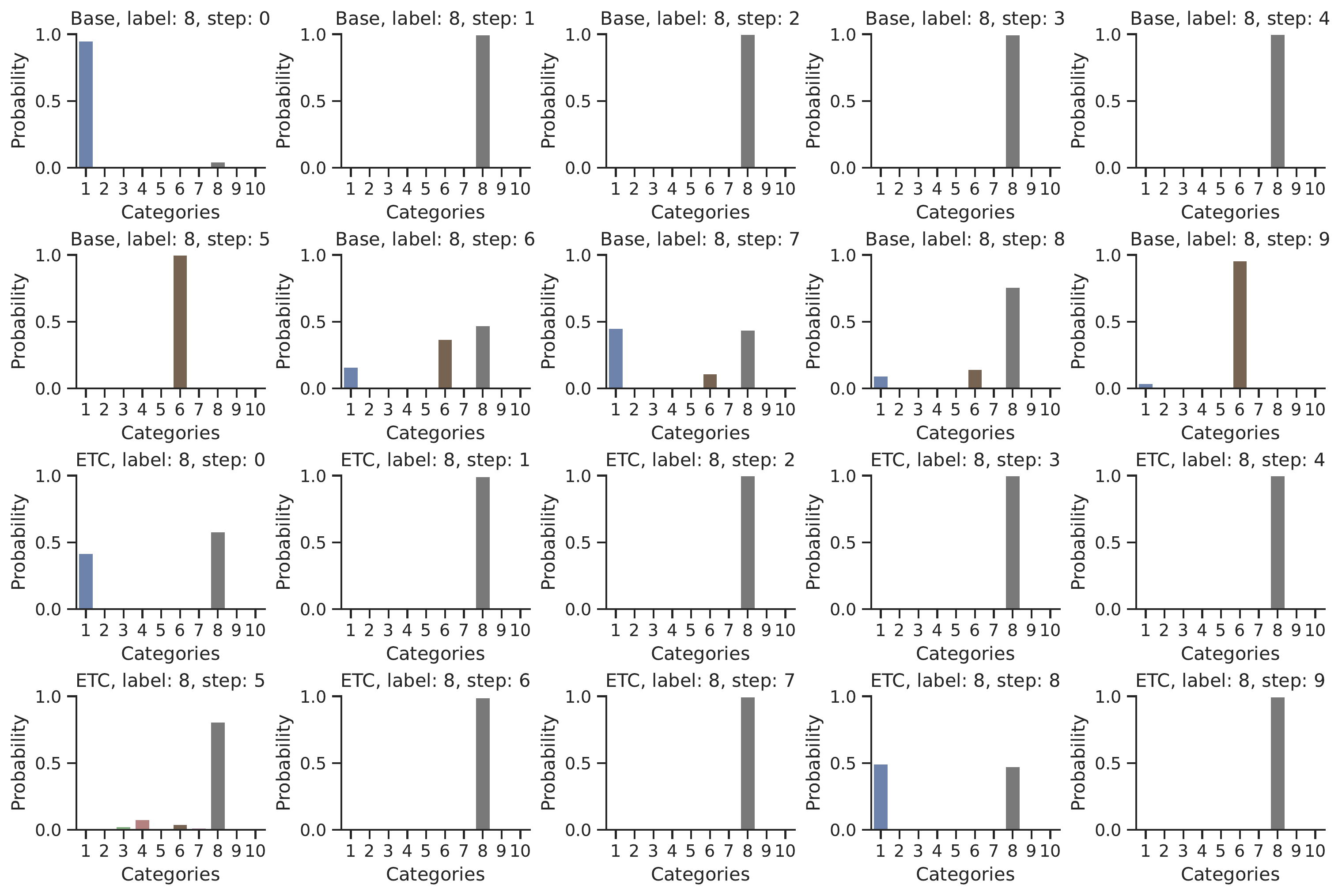}
\caption{The comparison between the base method and our method on output distribution at different timesteps for the sample in DVS-CIAFR10}
\label{distri}
 \end{figure}

Meanwhile, we have visualized the outputs at different timesteps in the final layer. As shown in Fig.~\ref{distri}, we visualize the distribution of the outputs at different timesteps for  the samples in the DVS-CIFAR10 dataset. It can be observed that the output distributions of the traditional algorithm vary significantly at different timesteps. For instance, the lsample's output distribution is accurate in the early moments, but from the sixth moment onwards, the distribution begins to fluctuate. This led to a sample that should have been categorized as 8 being incorrectly predicted as 6, 8, 1, 8, 6. In contrast, the output distribution of the ETC algorithm is much more consistent. By enhancing the consistency of the temporal distribution, the ETC algorithm has dramatically improved the accuracy of the overall distribution of outputs at each moment. This allows us to achieve high-precision predictions in the testing phase with fewer simulation steps. This characteristic greatly facilitates the deployment of SNNs on various edge devices.
 
 \section{Conclusion}
Unlike artificial neural networks, spiking neural networks can receive and generate inputs and outputs at multiple moments. However, the inconsistency of output distributions at different moments often requires longer simulation steps to produce stable and precise outputs. This seriously affects the performance of spiking neural networks and increases their latency. In this study, we propose a strategy to enhance temporal consistency, aiming to reduce the inconsistency in the output distributions at different timesteps. The approach significantly improves the performance of spiking neural networks on multiple datasets and effectively reduces latency. We can achieve a high accuracy in the testing phase with only short simulation steps. In biological neurons, there exist efficient coding methods that adaptively encode diverse inputs, along with decoding methods that accurately interpret the output information from different moments~\cite{rolls2011neuronal,quiroga2013principles,avitan2018code}. In the future, we can take more inspiration from the biologically plausible encoding and decoding method to process the information more effectively. 
\section*{Acknowledgement}
This work was supported by the National Key Research and Development Program (Grant No. 2020AAA0104305), and the Strategic Priority Research Program of the Chinese Academy of Sciences (Grant No. XDB32070100).

\bibliography{refs}

\begin{thebibliography}{10}

\bibitem{roy2019towards}
Kaushik Roy, Akhilesh Jaiswal, and Priyadarshini Panda.
\newblock Towards spike-based machine intelligence with neuromorphic computing.
\newblock {\em Nature}, 575(7784):607--617, 2019.

\bibitem{zeng2022braincog}
Yi~Zeng, Dongcheng Zhao, Feifei Zhao, Guobin Shen, Yiting Dong, Enmeng Lu, Qian
  Zhang, Yinqian Sun, Qian Liang, Yuxuan Zhao, et~al.
\newblock Braincog: A spiking neural network based brain-inspired cognitive
  intelligence engine for brain-inspired ai and brain simulation.
\newblock {\em arXiv preprint arXiv:2207.08533}, 2022.

\bibitem{dong2022unsupervised}
Yiting Dong, Dongcheng Zhao, Yang Li, and Yi~Zeng.
\newblock An unsupervised spiking neural network inspired by biologically
  plausible learning rules and connections.
\newblock {\em arXiv preprint arXiv:2207.02727}, 2022.

\bibitem{zheng2021high}
Yajing Zheng, Lingxiao Zheng, Zhaofei Yu, Boxin Shi, Yonghong Tian, and Tiejun
  Huang.
\newblock High-speed image reconstruction through short-term plasticity for
  spiking cameras.
\newblock In {\em Proceedings of the IEEE/CVF Conference on Computer Vision and
  Pattern Recognition}, pages 6358--6367, 2021.

\bibitem{han2020rmp}
Bing Han, Gopalakrishnan Srinivasan, and Kaushik Roy.
\newblock Rmp-snn: Residual membrane potential neuron for enabling deeper
  high-accuracy and low-latency spiking neural network.
\newblock In {\em Proceedings of the IEEE/CVF conference on computer vision and
  pattern recognition}, pages 13558--13567, 2020.

\bibitem{wang2022signed}
Yuchen Wang, Malu Zhang, Yi~Chen, and Hong Qu.
\newblock Signed neuron with memory: Towards simple, accurate and
  high-efficient ann-snn conversion.
\newblock In {\em International Joint Conference on Artificial Intelligence},
  2022.

\bibitem{liu2022spikeconverter}
Fangxin Liu, Wenbo Zhao, Yongbiao Chen, Zongwu Wang, and Li~Jiang.
\newblock Spikeconverter: An efficient conversion framework zipping the gap
  between artificial neural networks and spiking neural networks.
\newblock In {\em Proceedings of the AAAI Conference on Artificial
  Intelligence}, volume~36, pages 1692--1701, 2022.

\bibitem{li2022efficient}
Yang Li and Yi~Zeng.
\newblock Efficient and accurate conversion of spiking neural network with
  burst spikes.
\newblock {\em arXiv preprint arXiv:2204.13271}, 2022.

\bibitem{guo2022loss}
Yufei Guo, Yuanpei Chen, Liwen Zhang, Xiaode Liu, Yinglei Wang, Xuhui Huang,
  and Zhe Ma.
\newblock Im-loss: information maximization loss for spiking neural networks.
\newblock {\em Advances in Neural Information Processing Systems}, 35:156--166,
  2022.

\bibitem{li2021differentiable}
Yuhang Li, Yufei Guo, Shanghang Zhang, Shikuang Deng, Yongqing Hai, and Shi Gu.
\newblock Differentiable spike: Rethinking gradient-descent for training
  spiking neural networks.
\newblock {\em Advances in Neural Information Processing Systems},
  34:23426--23439, 2021.

\bibitem{dengtemporal}
Shikuang Deng, Yuhang Li, Shanghang Zhang, and Shi Gu.
\newblock Temporal efficient training of spiking neural network via gradient
  re-weighting.
\newblock {\em arXiv preprint arXiv:2202.11946}, 2022.

\bibitem{duan2022temporal}
Chaoteng Duan, Jianhao Ding, Shiyan Chen, Zhaofei Yu, and Tiejun Huang.
\newblock Temporal effective batch normalization in spiking neural networks.
\newblock {\em Advances in Neural Information Processing Systems},
  35:34377--34390, 2022.

\bibitem{fang2021incorporating}
Wei Fang, Zhaofei Yu, Yanqi Chen, Timoth{\'e}e Masquelier, Tiejun Huang, and
  Yonghong Tian.
\newblock Incorporating learnable membrane time constant to enhance learning of
  spiking neural networks.
\newblock In {\em Proceedings of the IEEE/CVF International Conference on
  Computer Vision}, pages 2661--2671, 2021.

\bibitem{dingbiologically}
Jianchuan Ding, Bo~Dong, Felix Heide, Yufei Ding, Yunduo Zhou, Baocai Yin, and
  Xin Yang.
\newblock Biologically inspired dynamic thresholds for spiking neural networks.
\newblock In {\em Advances in Neural Information Processing Systems}.

\bibitem{yaoglif}
Xingting Yao, Fanrong Li, Zitao Mo, and Jian Cheng.
\newblock Glif: A unified gated leaky integrate-and-fire neuron for spiking
  neural networks.
\newblock In {\em Advances in Neural Information Processing Systems}.

\bibitem{yin2021accurate}
Bojian Yin, Federico Corradi, and Sander~M Boht{\'e}.
\newblock Accurate and efficient time-domain classification with adaptive
  spiking recurrent neural networks.
\newblock {\em Nature Machine Intelligence}, 3(10):905--913, 2021.

\bibitem{wang2022ltmd}
Siqi Wang, Tee~Hiang Cheng, and Meng-Hiot Lim.
\newblock Ltmd: Learning improvement of spiking neural networks with learnable
  thresholding neurons and moderate dropout.
\newblock {\em Advances in Neural Information Processing Systems},
  35:28350--28362, 2022.

\bibitem{wu2019direct}
Yujie Wu, Lei Deng, Guoqi Li, Jun Zhu, Yuan Xie, and Luping Shi.
\newblock Direct training for spiking neural networks: Faster, larger, better.
\newblock In {\em Proceedings of the AAAI conference on artificial
  intelligence}, volume~33, pages 1311--1318, 2019.

\bibitem{zheng2021going}
Hanle Zheng, Yujie Wu, Lei Deng, Yifan Hu, and Guoqi Li.
\newblock Going deeper with directly-trained larger spiking neural networks.
\newblock In {\em Proceedings of the AAAI Conference on Artificial
  Intelligence}, volume~35, pages 11062--11070, 2021.

\bibitem{yao2021temporal}
Man Yao, Huanhuan Gao, Guangshe Zhao, Dingheng Wang, Yihan Lin, Zhaoxu Yang,
  and Guoqi Li.
\newblock Temporal-wise attention spiking neural networks for event streams
  classification.
\newblock In {\em Proceedings of the IEEE/CVF International Conference on
  Computer Vision}, pages 10221--10230, 2021.

\bibitem{zhu2022tcja}
Rui-Jie Zhu, Qihang Zhao, Tianjing Zhang, Haoyu Deng, Yule Duan, Malu Zhang,
  and Liang-Jian Deng.
\newblock Tcja-snn: Temporal-channel joint attention for spiking neural
  networks.
\newblock {\em arXiv preprint arXiv:2206.10177}, 2022.

\bibitem{yao2023attention}
Man Yao, Guangshe Zhao, Hengyu Zhang, Yifan Hu, Lei Deng, Yonghong Tian, Bo~Xu,
  and Guoqi Li.
\newblock Attention spiking neural networks.
\newblock {\em IEEE Transactions on Pattern Analysis and Machine Intelligence},
  2023.

\bibitem{cheng2020lisnn}
Xiang Cheng, Yunzhe Hao, Jiaming Xu, and Bo~Xu.
\newblock Lisnn: Improving spiking neural networks with lateral interactions
  for robust object recognition.
\newblock In {\em IJCAI}, pages 1519--1525, 2020.

\bibitem{zhao2022backeisnn}
Dongcheng Zhao, Yi~Zeng, and Yang Li.
\newblock Backeisnn: A deep spiking neural network with adaptive self-feedback
  and balanced excitatory--inhibitory neurons.
\newblock {\em Neural Networks}, 154:68--77, 2022.

\bibitem{fang2021deep}
Wei Fang, Zhaofei Yu, Yanqi Chen, Tiejun Huang, Timoth{\'e}e Masquelier, and
  Yonghong Tian.
\newblock Deep residual learning in spiking neural networks.
\newblock {\em Advances in Neural Information Processing Systems},
  34:21056--21069, 2021.

\bibitem{zhou2022spikformer}
Zhaokun Zhou, Yuesheng Zhu, Chao He, Yaowei Wang, Shuicheng Yan, Yonghong Tian,
  and Li~Yuan.
\newblock Spikformer: When spiking neural network meets transformer.
\newblock {\em arXiv preprint arXiv:2209.15425}, 2022.

\bibitem{na2022autosnn}
Byunggook Na, Jisoo Mok, Seongsik Park, Dongjin Lee, Hyeokjun Choe, and Sungroh
  Yoon.
\newblock Autosnn: towards energy-efficient spiking neural networks.
\newblock In {\em International Conference on Machine Learning}, pages
  16253--16269. PMLR, 2022.

\bibitem{kim2022neural}
Youngeun Kim, Yuhang Li, Hyoungseob Park, Yeshwanth Venkatesha, and
  Priyadarshini Panda.
\newblock Neural architecture search for spiking neural networks.
\newblock In {\em Computer Vision--ECCV 2022: 17th European Conference, Tel
  Aviv, Israel, October 23--27, 2022, Proceedings, Part XXIV}, pages 36--56.
  Springer, 2022.

\bibitem{hinton2015distilling}
Geoffrey Hinton, Oriol Vinyals, and Jeff Dean.
\newblock Distilling the knowledge in a neural network.
\newblock {\em arXiv preprint arXiv:1503.02531}, 2015.

\bibitem{krizhevsky2009learning}
Alex Krizhevsky, Geoffrey Hinton, et~al.
\newblock Learning multiple layers of features from tiny images.
\newblock 2009.

\bibitem{xu2015empirical}
Bing Xu, Naiyan Wang, Tianqi Chen, and Mu~Li.
\newblock Empirical evaluation of rectified activations in convolutional
  network.
\newblock {\em arXiv preprint arXiv:1505.00853}, 2015.

\bibitem{russakovsky2015imagenet}
Olga Russakovsky, Jia Deng, Hao Su, Jonathan Krause, Sanjeev Satheesh, Sean Ma,
  Zhiheng Huang, Andrej Karpathy, Aditya Khosla, Michael Bernstein, et~al.
\newblock Imagenet large scale visual recognition challenge.
\newblock {\em International journal of computer vision}, 115(3):211--252,
  2015.

\bibitem{li2017cifar10}
Hongmin Li, Hanchao Liu, Xiangyang Ji, Guoqi Li, and Luping Shi.
\newblock Cifar10-dvs: an event-stream dataset for object classification.
\newblock {\em Frontiers in neuroscience}, 11:309, 2017.

\bibitem{orchard2015converting}
Garrick Orchard, Ajinkya Jayawant, Gregory~K Cohen, and Nitish Thakor.
\newblock Converting static image datasets to spiking neuromorphic datasets
  using saccades.
\newblock {\em Frontiers in neuroscience}, 9:437, 2015.

\bibitem{bu2021optimal}
Tong Bu, Wei Fang, Jianhao Ding, PengLin Dai, Zhaofei Yu, and Tiejun Huang.
\newblock Optimal ann-snn conversion for high-accuracy and ultra-low-latency
  spiking neural networks.
\newblock In {\em International Conference on Learning Representations}, 2021.

\bibitem{rathi2019enabling}
Nitin Rathi, Gopalakrishnan Srinivasan, Priyadarshini Panda, and Kaushik Roy.
\newblock Enabling deep spiking neural networks with hybrid conversion and
  spike timing dependent backpropagation.
\newblock In {\em International Conference on Learning Representations}, 2019.

\bibitem{rathi2020diet}
Nitin Rathi and Kaushik Roy.
\newblock Diet-snn: Direct input encoding with leakage and threshold
  optimization in deep spiking neural networks.
\newblock {\em arXiv preprint arXiv:2008.03658}, 2020.

\bibitem{zhang2020temporal}
Wenrui Zhang and Peng Li.
\newblock Temporal spike sequence learning via backpropagation for deep spiking
  neural networks.
\newblock {\em Advances in Neural Information Processing Systems},
  33:12022--12033, 2020.

\bibitem{shen2022backpropagation}
Guobin Shen, Dongcheng Zhao, and Yi~Zeng.
\newblock Backpropagation with biologically plausible spatiotemporal adjustment
  for training deep spiking neural networks.
\newblock {\em Patterns}, page 100522, 2022.

\bibitem{dong2023temporal}
Yiting Dong, Dongcheng Zhao, and Yi~Zeng.
\newblock Temporal knowledge sharing enable spiking neural network learning
  from past and future.
\newblock {\em arXiv preprint arXiv:2304.06540}, 2023.

\bibitem{guo2022recdis}
Yufei Guo, Xinyi Tong, Yuanpei Chen, Liwen Zhang, Xiaode Liu, Zhe Ma, and Xuhui
  Huang.
\newblock Recdis-snn: Rectifying membrane potential distribution for directly
  training spiking neural networks.
\newblock In {\em Proceedings of the IEEE/CVF Conference on Computer Vision and
  Pattern Recognition}, pages 326--335, 2022.

\bibitem{wu2021liaf}
Zhenzhi Wu, Hehui Zhang, Yihan Lin, Guoqi Li, Meng Wang, and Ye~Tang.
\newblock Liaf-net: Leaky integrate and analog fire network for lightweight and
  efficient spatiotemporal information processing.
\newblock {\em IEEE Transactions on Neural Networks and Learning Systems},
  2021.

\bibitem{kugele2020efficient}
Alexander Kugele, Thomas Pfeil, Michael Pfeiffer, and Elisabetta Chicca.
\newblock Efficient processing of spatio-temporal data streams with spiking
  neural networks.
\newblock {\em Frontiers in Neuroscience}, 14:439, 2020.

\bibitem{ramesh2019dart}
Bharath Ramesh, Hong Yang, Garrick Orchard, Ngoc~Anh Le~Thi, Shihao Zhang, and
  Cheng Xiang.
\newblock Dart: distribution aware retinal transform for event-based cameras.
\newblock {\em IEEE transactions on pattern analysis and machine intelligence},
  42(11):2767--2780, 2019.

\bibitem{li2022neuromorphic}
Yuhang Li, Youngeun Kim, Hyoungseob Park, Tamar Geller, and Priyadarshini
  Panda.
\newblock Neuromorphic data augmentation for training spiking neural networks.
\newblock In {\em Computer Vision--ECCV 2022: 17th European Conference, Tel
  Aviv, Israel, October 23--27, 2022, Proceedings, Part VII}, pages 631--649.
  Springer, 2022.

\bibitem{shen2022eventmix}
Guobin Shen, Dongcheng Zhao, and Yi~Zeng.
\newblock Eventmix: An efficient augmentation strategy for event-based data.
\newblock {\em arXiv preprint arXiv:2205.12054}, 2022.

\bibitem{rolls2011neuronal}
Edmund~T Rolls and Alessandro Treves.
\newblock The neuronal encoding of information in the brain.
\newblock {\em Progress in neurobiology}, 95(3):448--490, 2011.

\bibitem{quiroga2013principles}
Rodrigo~Quian Quiroga and Stefano Panzeri.
\newblock {\em Principles of neural coding}.
\newblock CRC Press, 2013.

\bibitem{avitan2018code}
Lilach Avitan and Geoffrey~J Goodhill.
\newblock Code under construction: neural coding over development.
\newblock {\em Trends in Neurosciences}, 41(9):599--609, 2018.

\end{thebibliography}
\bibliographystyle{unsrt}

\end{document}